\documentclass[sigconf]{aamas} 

\usepackage{balance} 

\usepackage{algorithm}
\usepackage{algorithmic}
\usepackage{bm}

\usepackage{graphicx}
\usepackage{textcomp}
\usepackage{xcolor}
\usepackage{algorithm}
\usepackage{algorithmic}
\usepackage{bm}
\usepackage{subcaption}
\usepackage{adjustbox}

\setcopyright{ifaamas}
\acmConference[AAMAS '24]{Proc.\@ of the 23rd International Conference
on Autonomous Agents and Multiagent Systems (AAMAS 2024)}{May 6 -- 10, 2024}
{Auckland, New Zealand}{N.~Alechina, V.~Dignum, M.~Dastani, J.S.~Sichman (eds.)}
\copyrightyear{2024}
\acmYear{2024}
\acmDOI{}
\acmPrice{}
\acmISBN{}

\acmSubmissionID{<<478>>}

\title[AAMAS-2024 Formatting Instructions]{Optimal Task Assignment and Path Planning using Conflict-Based Search with Precedence and Temporal Constraints}

\author{Yu Quan Chong}
\affiliation{
  \institution{Carnegie Mellon University}
  \city{Pittsburgh, PA}
  \country{USA}}
\email{yuquanc@andrew.cmu.edu}

\author{Jiaoyang Li}
\affiliation{
  \institution{Carnegie Mellon University}
  \city{Pittsburgh, PA}
  \country{USA}}
\email{jiaoyangli@cmu.edu}

\author{Katia Sycara}
\affiliation{
  \institution{Carnegie Mellon University}
  \city{Pittsburgh, PA}
  \country{USA}}
\email{sycara@andrew.cmu.edu}

\begin{abstract}
The Multi-Agent Path Finding (MAPF) problem entails finding collision-free paths for a set of agents, guiding them from their start to goal locations. However, MAPF does not account for several practical task-related constraints. For example, agents may need to perform actions at goal locations with specific execution times, adhering to predetermined orders and timeframes. Moreover, goal assignments may not be predefined for agents, and the optimization objective may lack an explicit definition. To incorporate task assignment, path planning, and a user-defined objective into a coherent framework, this paper examines the Task Assignment and Path Finding with Precedence and Temporal Constraints (TAPF-PTC) problem. We augment Conflict-Based Search (CBS) to simultaneously generate task assignments and collision-free paths that adhere to precedence and temporal constraints, maximizing an objective quantified by the return from a user-defined reward function in reinforcement learning (RL). Experimentally, we demonstrate that our algorithm, CBS-TA-PTC, can solve highly challenging bomb-defusing tasks with precedence and temporal constraints efficiently relative to MARL and adapted Target Assignment and Path Finding (TAPF) methods.
\end{abstract}


\newcommand{\BibTeX}{\rm B\kern-.05em{\sc i\kern-.025em b}\kern-.08em\TeX}

\begin{document}


\pagestyle{fancy}
\fancyhead{}


\maketitle 


\section{Introduction}
\label{sec:introduction}

Multi-Agent Path Finding (MAPF) is a growing field with applications in fields such as transportation and logistics \cite{stern2019multi, wurman2008coordinating, morris2016planning}. Most MAPF algorithms applied in cases like warehouse pick-and-place operations generate collision-free plans to bring agents from their start to goal locations. However, in real-world scenarios, agents often have interdependent tasks that require actions to be performed in a specific order (precedence constraints) and/or within a specific timeframe (temporal constraints) at specific locations. For example, agents in manufacturing plants need to perform various interdependent tasks such as welding, painting, and assembly in a specific order and within certain timeframes to avoid product defects. Existing MAPF algorithms \cite{ferner2013odrm, goldenberg2014enhanced, wagner2015subdimensional, silver2005cooperative, ma2019searching, sharon2015conflict, shaw1998using} do not incorporate both precedence and temporal constraints when generating paths as well as the execution time of actions at goal locations. 

Furthermore, in many real-world problems, the underlying goal is to maximize a user-defined objective instead of minimizing the makespan or sum of path costs, which are usually of lower priority. The notion of a user-defined objective can be quantified through the reinforcement learning (RL) framework via returns from a user-defined reward function. For example, a sparse reward function would only give a positive reward (e.g. $+1$) when a task is completed and a reward of zero otherwise. Hence, generating an optimal task assignment that maximizes the user-defined objective that is typically unknown is also vital. In addition, task assignment and path planning are highly coupled as certain task assignments could lead to fewer conflicts during path planning, increasing the probability of finding an optimal solution. 

In general, many multi-agent problems based on real-world scenarios can be theoretically framed as multi-agent reinforcement learning (MARL) problems with a user-defined reward function. However, using MARL methods to train agents to learn to solve tasks involving path planning presents itself with many challenges such as sample efficiency. For example, PRIMAL \cite{sartoretti2019primal}, a framework that combines RL with imitation learning (IL) from an expert centralized MAPF planner like CBS~\cite{sharon2015conflict} and ODrM*~\cite{wagner2015subdimensional} to solve MAPF problems, had a training time of 20 days. Hence, rather than training agents to learn the path planning and additional skills necessary to solve tasks from scratch, it would be ideal to devise a single coherent framework that can incorporate task assignment, path planning, and a user-defined objective to solve tasks. 

Therefore, we examine the Task Assignment and Path Finding with Precedence and Temporal Constraints (TAPF-PTC) problem, where the goal is to simultaneously generate task assignments with feasible and collision-free paths for agents that adhere to precedence and temporal constraints to maximize a user-defined objective and present the Conflict-Based Search with Task Assignment, Precedence and Temporal Constraints (CBS-TA-PTC) algorithm. Experimental results demonstrate that CBS-TA-PTC can solve highly challenging bomb-defusing tasks with precedence and temporal constraints efficiently relative to MARL and adapted Target Assignment and Path Finding (TAPF)~\cite{ma2016optimal} methods.

\section{Background}
\label{sec:background}

In this section, we provide a brief overview of MAPF, CBS, MARL, and TAPF and highlight various related works to motivate and provide the necessary background for TAPF-PTC and CBS-TA-PTC.

\subsection{Multi-Agent Path Finding (MAPF)}
\label{subsec:background_mapf}

MAPF is characterised by an undirected graph $G = (V, E)$ and a set of $m$ agents $\{a_{1}, \dots, a_{m}\}$, where each agent $a_{i}$ possesses a start vertex $s_{i} \in V$ and goal vertex $g_{i} \in V$. Each agent can either move to a neighboring vertex or remain at its current vertex at unit cost. An agent's path is a sequence of actions that brings it from its start to its goal vertex and the path cost is the combined cost of all actions in the path. We consider two types of conflicts: 1) Vertex conflict $\langle a_{i}, a_{j}, v, t\rangle$ iff $a_{i}$ and $a_{j}$ are at the same vertex $v$ at the same timestep $t$, 2) Edge conflict $\langle a_{i}, a_{j}, u, v, t\rangle$ iff $a_{i}$ and $a_{j}$ traverse the same edge $\langle u, v \rangle$ in opposite directions between timesteps $t$ and $t + 1$. A constraint is a tuple $\langle a_{i}, v, t\rangle$ where $a_{i}$ is prohibited from occupying $v$ at timestep $t$. A solution to MAPF is a set of conflict-free paths for all agents and an optimal solution is a solution that gives the minimum for an objective function, which is typically the sum of path costs (sum of the path costs of all agents' paths) or the makespan (maximum path cost of all agents' paths). Finding an optimal solution for MAPF is known to be an NP-hard problem even when approximating optimal solutions \cite{lavalle2006planning, ma2016multi, ma2019searching}. 

\subsection{Conflict-Based Search (CBS)}
\label{subsec:background_cbs}

CBS \cite{sharon2015conflict} is a complete and optimal two-level MAPF algorithm. The high-level search is a best-first search with respect to the cost of a node about a binary Constraint Tree (CT), where each node contains a set of path constraints for various agents, a set of paths for each agent consistent with the path constraints, and a cost depending on the objective of the MAPF problem. CBS starts with the root CT node with an empty set of constraints and generates paths for each agent computed using the low-level planner. The high-level search evaluates the set of paths of each agent for conflicts, where a conflict-free CT node is returned as the goal node. Otherwise, it selects a conflict between the paths of two conflicting agents, splits the CT node into two child CT nodes, and appends a new constraint to each child CT node for each agent. The new constraints forbid either conflicting agents from utilizing the conflicting edge or vertex at the conflicting timestep. The low-level planner then recomputes the path of the constrained agent in the child CT nodes to be consistent with the new constraint set.

\subsection{Multi-Agent Reinforcement Learning (MARL)}
\label{subsec:background_marl}

MARL is a branch of RL whose objective is to train agents who can learn to cooperate or compete with each other in complex environments to achieve a common or individual goal. In partially observable Markov game \cite{littman1994markov} of $N$ agents, the set of states, actions and observations are given by $\mathcal{S}$, $\mathcal{A}_{1}, \dots, \mathcal{A}_{N}$ and $\mathcal{O}_{1}, \dots, \mathcal{O}_{N}$ respectively for each agent. A distribution $\rho: \mathcal{S} \xrightarrow[]{} [0, 1]$ selects the initial states. Each agent obtains an observation, $o_{i}: \mathcal{S} \xrightarrow[]{} \mathcal{O}_{i}$, follows a policy, $\bm{\pi}_{\theta_{i}}: \mathcal{O}_{i} \times \mathcal{A}_{i} \xrightarrow[]{} [0, 1]$ that generates the next state following the state transition function, $\mathcal{T}: \mathcal{S} \times \mathcal{A}_{1} \times \dots \times \mathcal{A}_{N} \xrightarrow[]{} \mathcal{S}$ and receives a reward, $r_{i}: \mathcal{S} \times \mathcal{A}_{1} \times \dots \times \mathcal{A}_{N} \xrightarrow[]{} \mathbb{R}$. Each agent aims to maximize its expected return $\mathbb{E}(G_{i}^{t})$, where return $G_{i}^{t} = \sum^{T_{term}}_{k=0} \gamma^{k} r_{i, t + k + 1}$, is the sum of discounted rewards from timestep $t$ and $\gamma$ and $T_{term}$ are the discount rate and the terminal timestep respectively.

\subsection{Target Assignment and Path Finding (TAPF)}
\label{subsec:background_tapf}

TAPF is characterised by an undirected graph $G = (V, E)$ with $K$ teams $\{team_{1}, \dots, team_{K}\}$, where each team $team_{i}$ consist of $K_{i}$ agents $\{a^{i}_{1}, \dots, a^{i}_{K_{i}}\}$ and are given unique target goal vertices $\{g^{i}_{1}, \dots, g^{i}_{K_{i}}\}$. Each agent $a^{i}_{j}$ has a unique start vertex $s^{i}_{j}$ and must move to a unique target goal vertex $g^{i}_{j'}$. Agents in $team_{i}$ are assigned to targets through a permutation based one-to-one mapping $\phi$, i.e. $g^{i}_{j'} = \phi^{i}(a^{i}_{j})$. 
Beyond collisions between agents within a team, TAPF considers collisions between agents of different teams, i.e. between agent $a^{i}_{j}$ in $team_{i}$ and agent $a^{i'}_{j'}$ in $team_{i'}$. The team conflicts are adapted from MAPF as follows: 1) Vertex conflict $\langle team_{i}, team_{i'}, v, t\rangle$ iff $a^{i}_{j}$ and $a^{i'}_{j'}$ are at the same vertex $v$ at the same timestep $t$, 2) Edge conflict $\langle team_{i}, team_{i'}, u, v, t\rangle$ iff $a^{i}_{j}$ and $a^{i'}_{j'}$ traverse the same edge $\langle u, v \rangle$ in opposite directions between timesteps $t$ and $t + 1$.

\subsection{Related Work}
\label{sec:background_related_work}

CBS with Precedence Constraints (CBS-PC)~\cite{zhang2022multi}, is a complete and optimal algorithm for MAPF but fails to consider temporal constraints between goals. CBS with Task Assignment (CBS-TA)~\cite{honig2018conflict} and the Conflict-Based Min-Cost-Flow (CBM) algorithm~\cite{ma2016optimal} are complete and optimal for TAPF but fail to consider precedence and temporal constraints between goals. As enumerating all task assignments is unfeasible combinatorics-wise, CBS-TA~\cite{honig2018conflict} generates assignments on demand using any optimal assignment algorithm (e.g. Hungarian method~\cite{kuhn1955hungarian}) to find the optimal solution. Using target swapping from an initial arbitrary target assignment, TSWAP is complete and highly scalable but sub-optimal for TAPF~\cite{okumura2023solving}. Another framework uses a marginal-cost assignment heuristic and a meta-heuristic improvement strategy based on Large Neighbourhood Search~\cite{shaw1998using} to minimize total travel delay~\cite{chen2021integrated}. However, the minimization of metrics such as the sum of path costs or makespan need not maximize (and can even reduce) the return in general. Hence, their frameworks cannot be leveraged. In addition, all of the above fails to consider the execution time of actions. 

\section{Problem Definition}
\label{sec:problem_definition}

TAPF-PTC is characterised by an undirected graph $G = (V, E)$, a set of $N$ agents $\{a_{1}, \dots, a_{N}\}$, and a task $T$, which is a set of $M$ goals $\{g^{1}, \dots, g^{M}\}$ with given temporal and precedence constraints. Each agent has an initial state $s_{i} \in \mathcal{S}$ consisting of a start vertex $v_{i} \in V$, where $v_{i} \in s_{i}$. Each goal $g^{i}$ is a tuple comprising a goal action $g^{i}.act \in \mathcal{A} \backslash \mathcal{A}_{v}$ and a goal vertex $g^{i}.v \in V$, where $\mathcal{A}$ and $\mathcal{A}_{v}$ are the set of all actions and actions for vertex movements, respectively. We need to find a binary $N \times M$ matrix $E$ that assigns goals to agents via a many-to-one mapping such that each agent $a_{i}$ has a sequence of $l_{i}$ goals $[g^{1}_{i}, \dots, g^{l_{i}}_{i}]$, where $\sum^{N}_{i} l_{i} = M$. An agent $a_{i}$ reaches $g^{j}_{i}.v$ and performs $g^{j}_{i}.act$ for $\alpha(g^{j})$ timesteps while waiting there. We further use 
$\mu(g^{j}_{i})$ and $\tau(g^{j}_{i})$ to denote the timestep when agent $a_i$ start and finish performing action $g^{j}_{i}.act$ at vertex $g^{j}_{i}.v$, respectively, so $\tau(g^{j}_{i}) - \mu(g^{j}_{i}) = \alpha(g^{j}_{i})$. A path segment for an agent $a_{i}$ consists of the sequence of states and actions from the completion of $g^{j - 1}_{i}$ to the completion of $g^{j}_{i}$, with the full path being the sequential concatenation of the path segments of $l_{i}$ goals. We introduce two additional temporal constraints: 1) Absolute temporal range constraints $\mathcal{T}_{abs}$ and 2) Inter-goal temporal range constraints $\mathcal{T}_{inter}$, in addition to the precedence constraints $\mathcal{T}_{prec}$ in CBS-PC \cite{zhang2022multi} and the vertex/edge constraints in CBS \cite{sharon2015conflict}. A solution to TAPF-PTC consists of a goal assignment and a set of conflict-free paths satisfying the above constraints.

\subsection{Absolute Temporal Range Constraints}
\label{subsec:problem_definition_absolute_temporal_range_constraint}

An absolute temporal range constraint $\langle t^{g^{j}_{i}}_{lower}, t^{g^{j}_{i}}_{upper}, \gamma \rangle \in \mathcal{T}_{abs}$ is a tuple where $t^{g^{j}_{i}}_{lower}$ and $t^{g^{j}_{i}}_{upper}$ represent lower and upper bounds of the range of timesteps where either the execution or completion timestep, denoted by $\gamma \in [\tau, \mu]$, for agent $a_{i}$ for $g^{j}_{i}$ must fall within, i.e., $t^{g^{j}_{i}}_{lower} \leq \gamma(g^{j}_{i}) \leq t^{g^{j}_{i}}_{upper}$. It allows the user to define the temporal range where $g^{j}_{i}$ must start execution or be completed. Using completion timestep as an example, an absolute temporal range conflict occurs when there exists $g^{j}_{i}$ with the absolute temporal range constraint $\langle t^{g^{j}_{i}}_{lower}, t^{g^{j}_{i}}_{upper}, \tau \rangle \in \mathcal{T}_{abs}$ and the following occurs:$\tau(g^{j}_{i}) < t^{g^{j}_{i}}_{lower}$ or $\tau(g^{j}_{i}) > t^{g^{j}_{i}}_{upper}$. A motivating example in the real world could be in the context of a restaurant with automated food service using robot waiters, where absolute temporal range constraints can be used to enforce the time in which the ordered items must be served to customers from the time of their order for good customer service. 

\subsection{Inter-Goal Temporal Constraints}
\label{subsec:problem_definition_inter_goal_temporal_constraint}

An inter-goal temporal constraint is a tuple consisting of two goals $g^{j}_{i}$ and $g^{j'}_{i'}$ with a temporal constraint between their respective execution or completion timesteps. For example, the following relation: $\tau(g^{j'}_{i'}) - \tau(g^{j}_{i}) \leq t_{inter}$ is denoted by $\langle \tau(g^{j}_{i}), \tau(g^{j'}_{i'}), t_{inter} \rangle \in \mathcal{T}_{inter}$. It allows the user to define the temporal range where $g^{j'}_{i'}.act$ must be executed or completed relative to the execution or completion of $g^{j}_{i}.act$. Considering completion timesteps for both $g^{j}_{i}$ and $g^{j'}_{i'}$ as an example, an inter-goal temporal conflict occurs with inter-goal temporal constraint $\langle \tau(g^{j}_{i}), \tau(g^{j'}_{i'}), t_{inter} \rangle \in \mathcal{T}_{inter}$ when the following relation $\tau(g^{j'}_{i'}) - \tau(g^{j}_{i}) > t_{inter}$ is true. Returning to the automated food service context, inter-goal temporal constraints can be used to ensure that orders for customers dining together must all arrive within a specified timeframe upon the arrival of the first item in the order. This would ensure that customers dining together would not have to wait for long for each other meals to arrive to have their meals together as part of good customer service.

\subsection{Precedence Constraints}
\label{subsec:problem_definition_precedence_constraint}

A precedence constraint is a tuple consisting of two goals $g^{j}_{i}$ and $g^{j'}_{i'}$ where their respective execution or completion timesteps are constrained precedence-wise. For example, $\langle \tau(g^{j}_{i}), \mu(g^{j'}_{i'}) \rangle \in \mathcal{T}_{prec}$ asserts the following relation: $\tau(g^{j}_{i}) < \mu(g^{j'}_{i'})$, i.e. the execution of $g^{j'}_{i'}.act$ must occur after the completion of $g^{j}_{i}.act$. It allows the user to define the order in which actions at goal vertices must be executed or completed relative to the execution or completion of other actions, which expands beyond \cite{zhang2022multi} that addresses precedence constraints for completion timesteps only, i.e. $\langle \tau(g^{j}_{i}), \tau(g^{j'}_{i'}) \rangle \in \mathcal{T}_{prec}$, with zero execution time. This is insufficient given many cases where it is necessary for an action to execute only after the completion of another action (e.g. assembling tasks). Considering completion timesteps for both $g^{j}_{i}$ and $g^{j'}_{i'}$ as an example, a precedence conflict occurs with precedence constraint $\langle \tau(g^{j}_{i}), \tau(g^{j'}_{i'}) \rangle \in \mathcal{T}_{prec}$ when the following relation $\tau(g^{j}_{i}) \geq \tau(g^{j'}_{i'})$ is true. Note that precedence constraints introduced here and in \cite{zhang2022multi} can be seen to be a special case of inter-goal temporal constraints defined in Section~\ref{subsec:problem_definition_inter_goal_temporal_constraint} given that $t_{inter}$ can be any real number. Returning to the automated food service context, precedence constraints can be used to ensure that items of a specific attribute are not served before others, e.g. desserts are not served before the main course. 

\subsection{Theoretical Overview}
\label{subsec:problem_definition_theoretical_overview}

The MAPF problem is a specific subclass of the TAPF-PTC problem where $N = M$, $E$ is a permutation matrix and the timestep required to execute the action is $0$, i.e. $\alpha(g^{j}_{i}) = 0$, $\mathcal{T}_{abs} = \mathcal{T}_{inter} = \mathcal{T}_{prec} = \emptyset$. Hence, finding an optimal solution with respect to makespan or sum of path costs for TAPF-PTC is also NP-Hard \cite{lavalle2006planning, ma2016multi, ma2019searching}. 

\section{CBS-TA-PTC}
\label{sec:cbs_ta_ptc}

\begin{algorithm}[t!]
    \caption{Task partition for CBS-TA-PTC}
    \label{alg:cbs_ta_ptc_task_partition}
    \begin{algorithmic}[1]
        \renewcommand{\algorithmicrequire}{\textbf{Input:}}
        \renewcommand{\algorithmicensure}{\textbf{Output:}}
        \REQUIRE Graph $G$, initial states $S$, task $T$, number of goals per $T_{sub}$ $\beta$
        \ENSURE Path for each agent for given $T$
        \STATE solution $\leftarrow \emptyset$
        \STATE $T$ $\leftarrow$ HeuristicSort($G$, $S$, $T$)
        \label{alg:cbs_ta_ptc_task_partition_heuristic_sort}
        \STATE $T$ $\leftarrow$ Partition($T$, $\beta$)
        \label{alg:cbs_ta_ptc_task_partition_partition}
        \FOR{$T_{sub}$ in $T$}
            \STATE $T_{sub}$ solution $\leftarrow$ CBS-TA-PTC($G$, $S$, $T_{sub}$, solution)
            \STATE solution.append($T_{sub}$ solution)
        \ENDFOR
        \RETURN solution
    \end{algorithmic}
\end{algorithm}

In this section, we introduce CBS-TA-PTC and provide an overview of its high-level conflict resolution and low-level path planning. In cases where $M$ is significantly larger than $N$, we partition $T$ into subtasks $\{T_{sub}\}$ using Algorithm~\ref{alg:cbs_ta_ptc_task_partition}, forming each $T_{sub}$ as a TAPF-PTC instance and solved by CBS-TA-PTC in Algorithm~\ref{alg:cbs_ta_ptc}. In Algorithm~\ref{alg:cbs_ta_ptc_task_partition}, a user-defined heuristic is used to sort the execution order of the goals for the task $T$ as shown in line~\ref{alg:cbs_ta_ptc_task_partition_heuristic_sort} in Algorithm~\ref{alg:cbs_ta_ptc_task_partition}. The ordered goals are then partitioned using a hyperparameter $\beta$, the number of goals per subtask, to form a subtask $T_{sub}$ as shown in line~\ref{alg:cbs_ta_ptc_task_partition_partition} to ensure computationally feasible task assignments.

\subsection{High-Level of CBS-TA-PTC}
\label{subsec:cbs_ta_ptc_high_level}

\begin{algorithm}[t!]
    \caption{CBS-TA-PTC()}
    \label{alg:cbs_ta_ptc}
    \begin{algorithmic}[1]
        \renewcommand{\algorithmicrequire}{\textbf{Input:}}
        \renewcommand{\algorithmicensure}{\textbf{Output:}}
        \REQUIRE Graph $G$, initial states $S$, subtask $T_{sub}$, past solutions
        \ENSURE Path for each agent for given $T_{sub}$
        \FOR{assignment in Combinations($T_{sub}$)} \label{alg:cbs_ta_ptc_combinations_start}
            \STATE R $\leftarrow$ new node
            \STATE R.constraints $\leftarrow \emptyset$
            \STATE R.assignment $\leftarrow$ assignment
            \STATE R.solution $\leftarrow$ Paths from LowLevel() \label{alg:cbs_ta_ptc_low_level_root_solution}
            \STATE R.cost $\leftarrow$ SumOfPathCost(R.solution)
            \STATE R.return $\leftarrow$ Return(S, R.solution, past solutions) \label{alg:cbs_ta_ptc_root_return}
            \STATE R.conflicts $\leftarrow$ Conflicts(R.solution)
            \IF{R.return is maximum return} \label{alg:cbs_ta_ptc_root_maximum_return_start}
                \RETURN R.solution
            \ENDIF \label{alg:cbs_ta_ptc_root_maximum_return_end}
            \STATE insert R to OPEN
        \ENDFOR \label{alg:cbs_ta_ptc_combinations_end}
        \WHILE{OPEN not empty} \label{alg:cbs_ta_ptc_while_open_start}
            \STATE P $\leftarrow$ pop node from OPEN with the highest return \label{alg:cbs_ta_ptc_open_highest_return}
            \IF{P.return is maximum return}
                \RETURN P.solution
            \ENDIF
            \STATE conflict $\leftarrow$ conflict in P w.r.t conflict resolution order \label{alg:cbs_ta_ptc_conflict_resolution_priority_order}
            \STATE constraints $\leftarrow$ ResolveConflict(conflict) \label{alg:cbs_ta_ptc_conflict_resolve_conflict}
            \FOR{constraint in constraints}
                \STATE Q $\leftarrow$ new node 
                \STATE Q.constraints $\leftarrow$ P.constraints.append(constraint)
                \IF{Q.constraints is unsolvable} \label{alg:cbs_ta_ptc_lp_solvable_start}
                    \STATE continue
                \ENDIF \label{alg:cbs_ta_ptc_lp_solvable_end}
                \STATE Q.assignment $\leftarrow$ P.assignment
                \STATE Q.solution $\leftarrow$ P.solution
                \STATE LowLevel() updates Q.solution for agent $a_{i}$ in constraint \label{alg:cbs_ta_ptc_low_level_update_solution} 
                \STATE Q.cost $\leftarrow$ SumOfPathCost(Q.solution)
                \STATE Q.return $\leftarrow$ Return(S, Q.solution, past solutions)
                \STATE Q.conflicts $\leftarrow$ Conflicts(Q.solution) 
                \STATE insert Q to OPEN
            \ENDFOR
        \ENDWHILE \label{alg:cbs_ta_ptc_while_open_end}
    \end{algorithmic}
\end{algorithm}

From lines~\ref{alg:cbs_ta_ptc_combinations_start}--\ref{alg:cbs_ta_ptc_combinations_end} in Algorithm~\ref{alg:cbs_ta_ptc}, the goals from the subtask are used to generate all possible combinations of assignments to agents, where each unique task assignment corresponds to a root CT node inserted to OPEN. An important property for root nodes would be the return as shown in line~\ref{alg:cbs_ta_ptc_root_return}, which is generated by evaluating the solution from the low level, with past solutions from previous subtasks, on an oracle that simulates the environment based on the user-defined reward function and dynamics. In practice, the oracle is typically implemented as a user-designed RL environment based on the Gymnasium API \cite{towers_gymnasium_2023}, where the solution from the low level with past solutions from previous subtasks can be rolled out as a single trajectory via the \texttt{reset} and \texttt{step} functions from the \texttt{Env} class in Gymnasium. The return can be simply calculated by summing up the rewards across timesteps and agents returned by the \texttt{step} function. Note that a solution with the maximum return implies a conflict-free solution by the design of the reward function (reward is positively correlated with conflict resolution).

CBS-TA-PTC uses the following conflict resolution priority ordering: 1) Absolute temporal range, 2) Precedence, 3) Inter-goal temporal, and 4) Vertex/Edge, to maximize the likelihood of generating a CT node with the maximum return. The natural first order of priority is to address the absolute temporal range conflicts as they are based on temporal ranges independent from other goals. Precedence conflicts take priority over inter-goal temporal conflicts as it is logical to ensure that actions are performed in the correct order before asserting that they are correctly spaced apart temporally. In addition, it is also likely resolving a precedence conflict would resolve any related inter-goal temporal conflict.

The following details the conflict resolution for inter-goal temporal conflicts using completion timesteps for all actions. When an inter-goal temporal constraint $\langle \tau(g^{j}_{i}), \tau(g^{j'}_{i'}), t_{inter} \rangle$ is violated, i.e. $\tau(g^{j'}_{i'}) - \tau(g^{j}_{i}) > t_{inter}$, CBS-TA-PTC generates two child CT nodes with one of the following inter-goal temporal completion timestep constraints added to either child CT nodes, letting $\tau(g^{j'}_{i'}) = t'$ at the current CT node:

\begin{enumerate}
    \item $\tau(g^{j}_{i}) \geq t' - t_{inter}$: Agent $a_{i}$ is to complete $g^{j}_{i}$ at or after timestep $t' - t_{inter}$. Agent $a_{i}$'s path in the corresponding child CT node is replanned and the inter-goal temporal conflict is resolved. \label{enum:inter_goal_temporal_constraint_1}
    \item $\tau(g^{j}_{i}) < t' - t_{inter}$: Agent $a_{i}$ is to complete $g^{j}_{i}$ before timestep $t' - t_{inter}$, which is already satisfied. Given the inter-goal temporal constraint, this forces agent $a_{i'}$ to complete $g^{j'}_{i'}$ at least one timestep earlier than before when replanning its path in the corresponding child CT node, i.e. $\tau(g^{j'}_{i'}) \leq t' - 1$. Hence, $\tau(g^{j'}_{i'})$ in the child CT node is guaranteed to decrease by one timestep. \label{enum:inter_goal_temporal_constraint_2}
\end{enumerate}

Agent $a_{i'}$ is unlikely to find a path following inter-goal temporal constraint (\ref{enum:inter_goal_temporal_constraint_2}) given that $t'$ is likely to be the earliest possible completion timestep for $g^{j'}_{i'}$ for agent $a_{i'}$. Hence, the child CT generated by it is likely to be pruned. For conciseness, absolute temporal range and precedence conflicts can be explained similarly. Line~\ref{alg:cbs_ta_ptc_while_open_start}--\ref{alg:cbs_ta_ptc_while_open_end} of Algorithm~\ref{alg:cbs_ta_ptc} highlights CBS-TA-PTC's conflict resolution process, where line~\ref{alg:cbs_ta_ptc_conflict_resolution_priority_order} generates conflicts based on the stated priority order and line~\ref{alg:cbs_ta_ptc_conflict_resolve_conflict} resolves the conflict by generating the necessary constraints to be inserted in child CT nodes to be pushed to OPEN. The solution node to be returned is the node with the maximum return as evaluated by the oracle. The solution for the entire task is then composed of the solutions of the subtask sequentially.

\subsection{Low-Level of CBS-TA-PTC}
\label{subsec:cbs_ta_ptc_low_level}

Multi-Label A* (MLA*) \cite{grenouilleau2019multi} is used to generate a minimum sum of path costs solution for a task assignment under the constraints on the CT node as shown in line~\ref{alg:cbs_ta_ptc_low_level_root_solution} and \ref{alg:cbs_ta_ptc_low_level_update_solution}. To speed up the low level, a linear programming module using one of the HiGHS solvers \cite{hallhighs, huangfu2018parallelizing} is implemented to prune CT nodes that are unsolvable given their constraints from the various types of conflicts before the MLA* search as shown in lines~\ref{alg:cbs_ta_ptc_lp_solvable_start}--\ref{alg:cbs_ta_ptc_lp_solvable_end} in Algorithm~\ref{alg:cbs_ta_ptc}.

\subsection{Theoretical Properties of CBS-TA-PTC}
\label{subsec:cbs_ta_ptc_theoretical_properties}

\begin{figure*}[t!]
  \centerline{\includegraphics[width=0.7\textwidth]{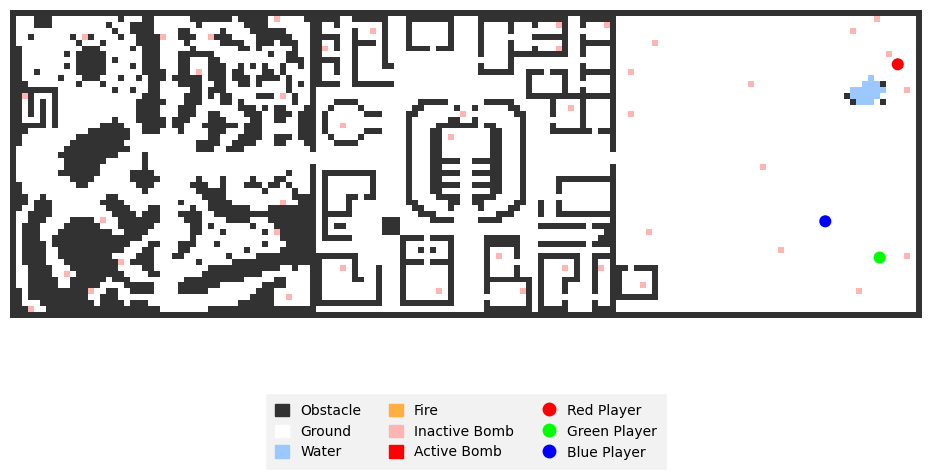}}
  \caption{An instance of the bomb-defusing task environment}
  \label{fig:gym_dragon}
\end{figure*}

By using user-defined heuristics to decompose the task into subtasks, CBS-TA-PTC is an incomplete and suboptimal algorithm. However, with the entire task as a single subtask, the following will show that CBS-TA-PTC is a complete and optimal algorithm and present its bounded sub-optimal solver. 

\subsubsection{Optimality of CBS-TA-PTC}
\label{subsubsec:cbs_ta_ptc_theoretical_properties_optimality}

CBS-TA-PTC enumerates all possible assignments and adds the next best possible task assignment with respect to the return to OPEN whenever a root CT node is expanded. Unlike the sum of path costs, given that the reward function is user-defined, the maximum return that an optimal solution gives is always known through the oracle. Naturally, the return of a CT node is a lower bound to the maximum return. It is proven in Lemma 2 in \cite{sharon2015conflict} that for each valid solution $p$, there exists at least one CT node $N$ in OPEN such that $N$ permits $p$, implying that there is at least one CT node in OPEN that permits the optimal solution at all times. CBS-TA-PTC expands CT nodes in a best-first manner with respect to return and the MLA* algorithm used for low level is optimal \cite{grenouilleau2019multi}. Hence, we can adapt Theorem 1 in \cite{sharon2015conflict} for CBS to show that CBS-TA-PTC returns an optimal solution. 

\subsubsection{Completeness of CBS-TA-PTC}
\label{subsubsec:cbs_ta_ptc_theoretical_properties_completeness}

Completeness includes the following claims: 1) CBS-TA-PTC will return a solution if one exists and 2) CBS-TA-PTC will identify an unsolvable problem. To prove the stated claims, we assume that every trajectory is a Markov game that terminates with a fixed terminal timestep, which is a reasonable assumption given that a significant number of tasks in the real world have practical deadlines. As a result, the set of possible constraints that can be added to CT nodes is finite. Hence, there is a finite number of CT nodes with a specific return for every return (a variant of Theorem 2 in \cite{sharon2015conflict}). As the MLA* used in the low level is complete \cite{grenouilleau2019multi}, Theorem 3 in \cite{sharon2015conflict} can then be easily adapted to show claim (1) that CBS-TA-PTC will return an optimal solution if one exists, given the optimality of CBS-TA-PTC proven in Section~\ref{subsubsec:cbs_ta_ptc_theoretical_properties_optimality}. Note that the return of every CT node is not monotonically non-decreasing, unlike the sum of path costs, as the resolution of a constraint may generate more conflicts that lead to a corresponding reduction in return. 

For claim (2), note that CBS on its own is unable to identify an unsolvable instance and requires an independent test by Yu and Rus~\cite{yu2015pebble} to determine the solvability of a given MAPF instance. This is the consequence of a lack of a terminal time horizon, resulting in an infinite set of constraints that can be added to CT nodes. A simple example is provided in~\cite{sharon2015conflict} of two agents attempting to switch locations, where the CT grows indefinitely through the addition of unlimited constraints and the algorithm does not terminate. Fortunately, given the initial assumption of a fixed terminal timestep and hence a finite set of possible constraints to be added to the CT, by exhaustive search over the CT and not finding a solution node with the maximum return, CBS-TA-PTC will be able to identify that the problem is unsolvable. 

\subsubsection{Bounded Sub-Optimal CBS-TA-PTC}
\label{subsubsec:cbs_ta_ptc_theoretical_properties_bounded_suboptimal}

A bounded sub-optimal variant of CBS-TA-PTC can be simply implemented by searching for a CT node with the maximum return multiplied by a suboptimality factor $\epsilon$.

\section{Experiments}
\label{sec:experiments}

The following sections detail the environment, experimental setup, and results. All code is implemented in Python.

\subsection{Environment}
\label{subsec:experiments_environment}

In the environment, three agents are tasked to defuse all the bombs on the map by a specific time limit. Each bomb has an ordered sequence of colors: red (R), green (G), or blue (B), with a sequence length ranging from $[1, 3]$, (e.g. RGB), with no repeated colors in a sequence. Each agent has wire-cutter tools with 2 out of the 3 colors that remove the matching color from the ordered sequence of a bomb. In addition, each bomb has a countdown timer (e.g. 15 seconds) that triggers/resets upon a correct defusing step, where the next sequence must be defused before the countdown. Each bomb also has a fuse timer (e.g. 3 minutes) that indicates the time when it must be fully defused. Furthermore, certain bombs have dependencies with other bombs, which dictates that the bomb it is dependent on must be fully defused or have exploded before any defusing steps on itself. Failure to adhere to the above constraints would result in the bomb exploding. A fully defused bomb gives a team reward proportionate to the sequence length of the bomb (e.g. RGB: $3 \times 10 = 30$). Figure~\ref{fig:gym_dragon} highlights the environment comprising three regions: Forest, Village, and Desert, derived from an implementation in a game called Minecraft \cite{minecraft}. A segmentation algorithm is implemented to obtain a graph representation of the map with approximately 40 nodes per region of the map that are generally equally spaced apart and represent an equal area of traversable grids. Note that a node can contain multiple bombs. The motivation for a graph representation instead of a grid world is to reduce the state space to aid learning for MARL methods. 

\subsection{Setup}
\label{subsec:experiments_setup}

\begin{table}[t!]
\centering
\caption{Environment Parameters}
\label{tab:environment_parameters}
\begin{tabular}{|c|c|}
\hline
\textbf{Parameters}          & \textbf{Values} \\ \hline
Mission Length (min)         & 15              \\ \hline
Seconds per Timestep         & [1, 3]          \\ \hline
Number of Bombs per Region   & [1, 15]         \\ \hline
Bomb Fuse Length (min)       & [1, 15]         \\ \hline
Bomb Dependency Chain Length & [1, 4]          \\ \hline
\end{tabular}
\end{table}

\subsubsection{Environment Setup}
\label{subsubsec:experiments_setup_environment}

Table~\ref{tab:environment_parameters} highlights the values used to instantiate the environment. The seconds per timestep parameter serves to convert real-time into timesteps to define the underlying kinematics and dynamics of the environment. For simplicity, node traversal and a defusing step take the same number of timesteps. All agents start at the same randomly selected location. Each bomb's location, sequence, and fuse length are randomly selected. A number of bombs per region are randomly selected from the range specified by the bomb dependency chain length parameter to form a bomb dependency chain. Note that a randomly generated environment instance may be unsolvable. To the best of our knowledge, there are no efficient optimal, and complete solvers for the TAPF-PTC problem to filter unsolvable cases. The only option available would be CBS-TA-PTC with the entire task as a subtask as elaborated in Section~\ref{subsec:cbs_ta_ptc_theoretical_properties}, which is prohibitively expensive to run combinatorics-wise with an increasing number of bombs. To increase the likelihood that a randomly generated environment is solvable, we allow for multiple agents to work on bombs on the same node at the same timestep. Hence, collisions between agents are not considered, which is not a key factor given the small number of agents. 

\subsubsection{Algorithm Setup}
\label{subsubsec:experiments_setup_algorithm}

CBS-TA-PTC parameters are as follows: 1) Number of bombs per subtask and 2) Task allocation heuristic. The number of bombs per subtask is a proxy parameter for the number of goals per subtask, $\beta$, in Algorithm~\ref{alg:cbs_ta_ptc_task_partition}, as the bomb sequence can be broken down into goals with defusing actions of a single color sequence. To keep the combinatorics of task allocation feasible, the number of bombs per subtask used will be in the range $[1, 3]$. The task allocation heuristic is the heuristic used for HeuristicSort() at line~\ref{alg:cbs_ta_ptc_task_partition_heuristic_sort} in Algorithm~\ref{alg:cbs_ta_ptc_task_partition}. We use the fuse length heuristic that sorts the bombs in ascending order of fuse length. For each combination of parameters, $10$ trials are executed with a timeout of $5$ minutes, and the success rate, optimality ratio, and runtime are recorded. Success is defined as finding an optimal solution before the timeout. The optimality ratio is the ratio between return and optimal return. The runtime includes only optimal solutions and all solutions (with cases due to timeout or no optimal solution found).

For comparisons with CBS-TA-PTC, we consider a state-of-the-art on-policy MARL algorithm, Multi-Agent Proximal Policy Optimisation (MAPPO)~\cite{yu2022surprising}, which has shown generally better performance in environments such as the StarCraft Multi-Agent Challenge (SMAC)~\cite{samvelyan19smac}, the Hanabi challenge~\cite{bard2020hanabi} and the multi-agent particle-world environment (MPE)~\cite{lowe2017multi, mordatch2017emergence} relative to various state-of-the-art off-policy MARL methods such as MADDPG~\cite{lowe2017multi} and QMIX~\cite{rashid2020monotonic}. The key motivation for considering MAPPO as a baseline is to evaluate whether the state-of-the-art MARL methods are capable of learning the necessary path planning and bomb-defusing skills to solve tasks in the given environment from scratch and their sample efficiency in terms of wall-clock training time. Such comparisons would ascertain if the underlying motivation behind our work is valid, that is MARL methods can be highly sample inefficient in training and an alternative solution that incorporates task assignment, path planning and a user-defined objective to solve tasks is more practical. MAPPO is trained in each region and the whole map, where each region has $15$ bombs without fuses and dependencies, a mission length of $15$ minutes, seconds per timestep parameter of $1$, and full observability over the environment for all agents. 

For comparisons with MAPF baselines, we implement a variant of CBS-TA \cite{honig2018conflict} that maximizes return under the same subtask assignment framework as CBS-TA-PTC highlighted in Algorithm~\ref{alg:cbs_ta_ptc_task_partition} but with vertex/edge constraints only. Under CBS-TA, upon the explosion of a bomb at a given timestep, CT nodes are generated in a naive manner with vertex constraints for each agent for that timestep, i.e. at least one agent is doing something wrong at that timestep. This variant of CBS-TA can be seen as CBS-TA-PTC without absolute temporal range, inter-goal temporal and precedence conflicts, It attempts to solve the TAPF-PTC problem by relying solely on vertex constraints to resolve the underlying precedence and temporal conflicts through a more exhaustive and inefficient best-first search through the CT with more CT nodes relative to CBS-TA-PTC. Nevertheless, it is worth highlighting that this variant of CBS-TA is optimal and complete like CBS-TA-PTC as elaborated in Section~\ref{subsec:cbs_ta_ptc_theoretical_properties}, under the assumption that the entire task is a single subtask.

\subsubsection{Hardware Setup}
\label{subsubsec:experiments_setup_hardware}

\begin{figure}[t!]
  \centerline{\includegraphics[width=\columnwidth]{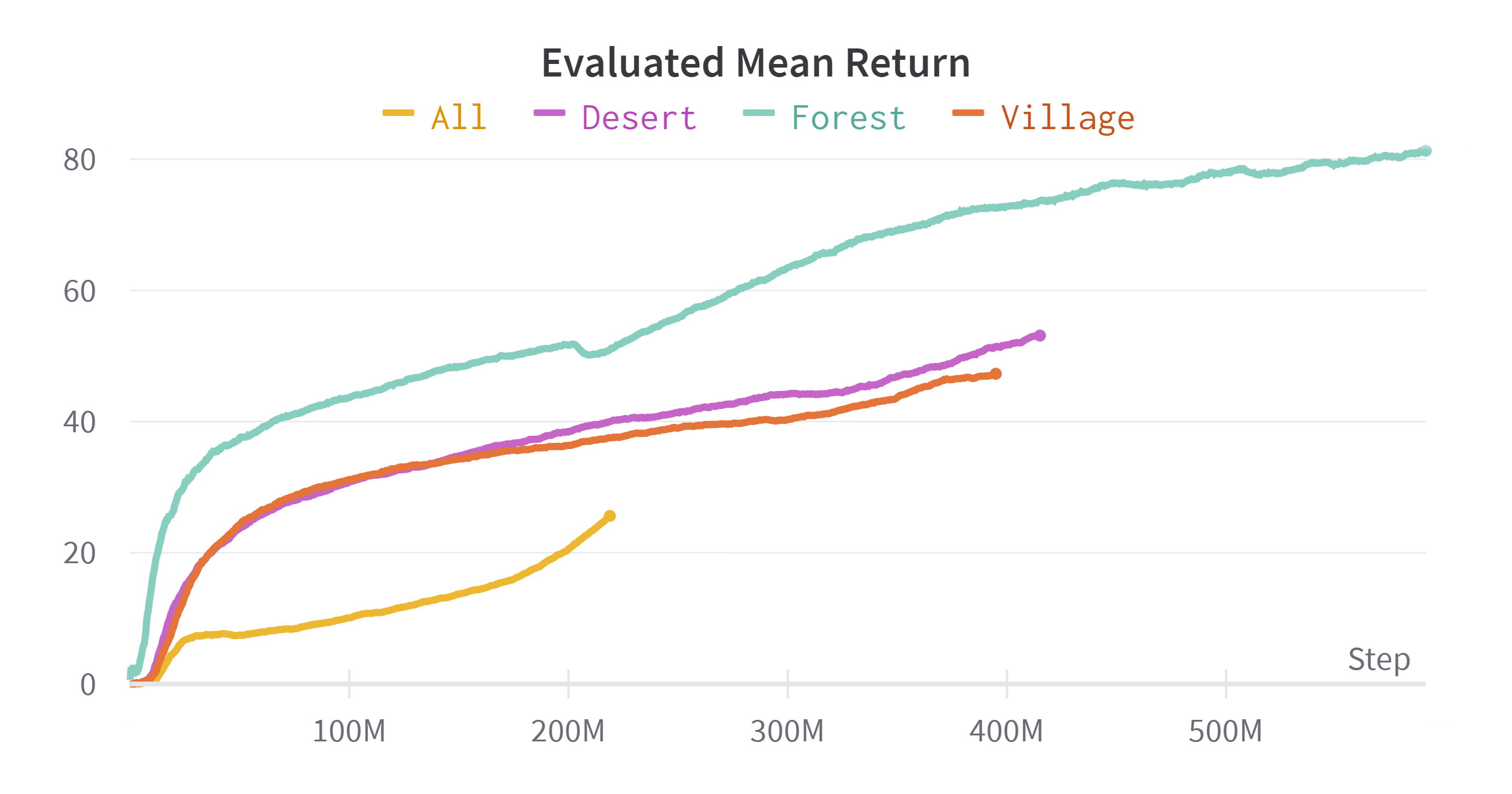}}
  \caption{Evaluated mean return of MAPPO from 16 rollouts}
  \label{fig:rmappo_gym_dragon_evaluated_mean_return}
\end{figure}

MAPPO is trained using a single Quadro RTX 6000 GPU. CBS-based algorithms are implemented on a laptop with the following specifications: 2.30GHz Intel Core i7-11800H with 16GB RAM.

\subsection{Results}
\label{subsec:experiments_results}

\begin{figure}[t!]
    \centering
    \subcaptionbox{Number of bombs per region \label{fig:plots_num_bombs_per_region}}{\includegraphics[width=\columnwidth,height=\columnwidth]{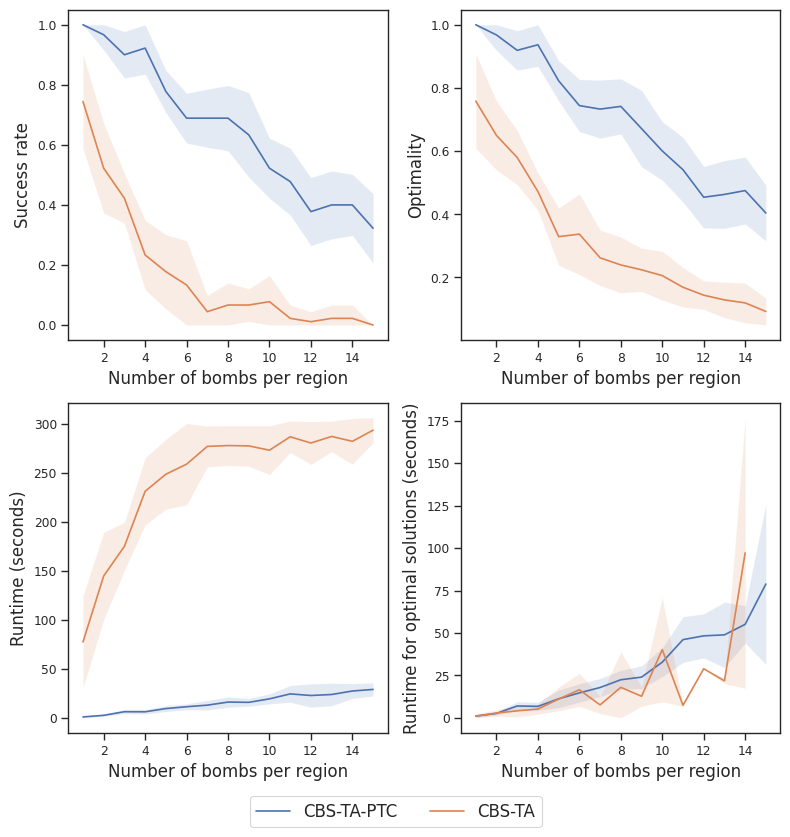}}
    \subcaptionbox{Number of bombs per subtask \label{fig:plots_num_bombs_per_subtask}}{\includegraphics[width=\columnwidth,height=\columnwidth]{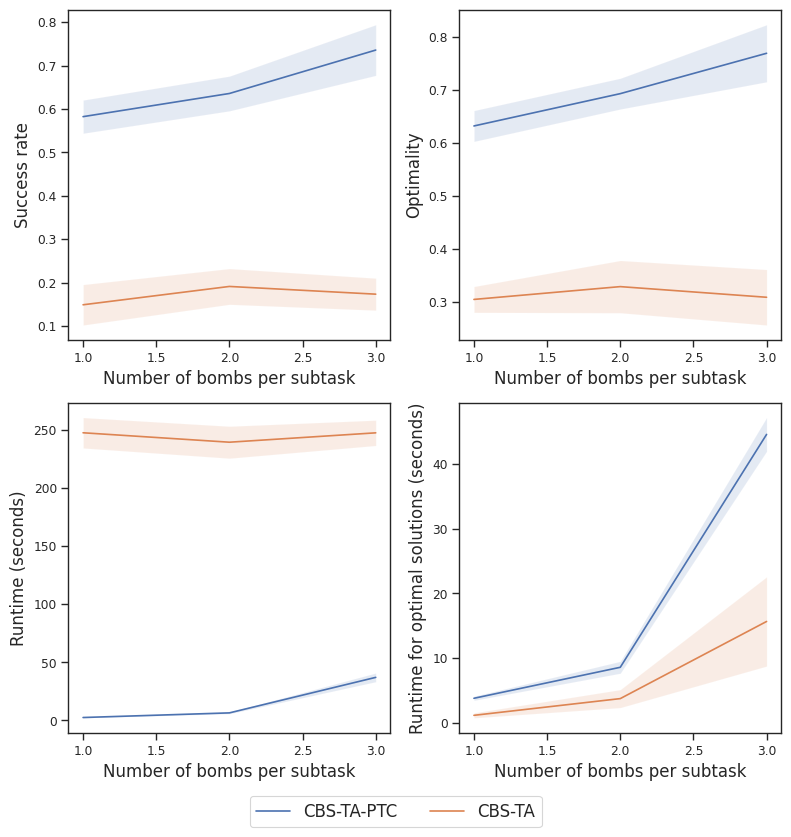}}
\end{figure}
\begin{figure}[t!]\ContinuedFloat
    \centering
    \subcaptionbox{Seconds per timestep \label{fig:plots_seconds_per_timestep}}{\includegraphics[width=\columnwidth,height=\columnwidth]{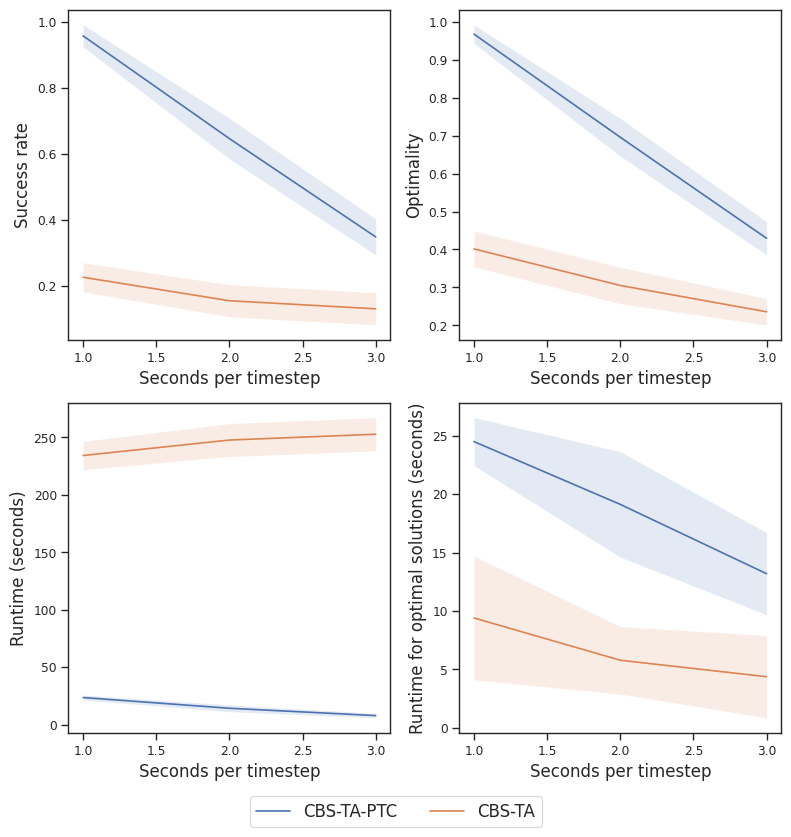}}
    \caption{Evaluated results for CBS-based algorithms, where the intervals highlight the standard deviation of averaged results across 10 trials.}
    \label{fig:cbs_results}
\end{figure}

\subsubsection{MAPPO}
\label{subsubsec:experiments_results_mappo}

Given $15$ bombs per region, the expected maximum return is $300$ per region. Figure~\ref{fig:rmappo_gym_dragon_evaluated_mean_return} highlights the evaluated mean return for MAPPO from $16$ evaluation rollouts. MAPPO is incapable of solving the bomb-defusing task optimally despite being given a significantly easier task without bomb fuses and dependencies as well as the smallest seconds per timestep parameter. In addition, the wall-clock training time for MAPPO was approximately two weeks for the stated results, highlighting the sample inefficiency of MARL methods.

\subsubsection{CBS}
\label{subsubsec:experiments_results_cbs}

Figure~\ref{fig:cbs_results} highlights the results obtained from the CBS-based algorithms averaged across all trials. CBS-TA-PTC is superior to CBS-TA across all parameters in general as CBS-TA-PTC always has a significantly higher success rate and optimality ratio compared to CBS-TA. While the success rate and optimality ratio naturally decrease with an increasing number of bombs per region for both CBS-TA-PTC and CBS-TA, the two metrics decay linearly for CBS-TA-PTC instead of exponentially for CBS-TA as shown in
Figure~\ref{fig:plots_num_bombs_per_region}. In addition, CBS-TA-PTC's runtime scales linearly with respect to the number of bombs per region in general with a runtime of approximately a minute for successful instances with the highest number of bombs per region, i.e. $15$. On the other hand, CBS-TA's runtime quickly plateaus towards the timeout of $5$ minutes, which is expected given its naive and inefficient manner of expanding the CT.

From Figure~\ref{fig:plots_num_bombs_per_subtask}, the success rate and optimality ratio improve linearly as the number of bombs per subtask increases for CBS-TA-PTC, which is expected as a larger subtask allows for CBS-TA-PTC to produce subtask solutions of higher quality by considering more constraints. Furthermore, with a smaller number of subtasks, given the fact that the solution for the entire task is composed of the solutions of the subtask sequentially, it is easier to find a good quality solution for a given subtask given a good initialization from past subtask solutions. On the other hand, CBS-TA does poorly in terms of success rate and optimality ratio regardless of the number of bombs per subtask given its naive and inefficient manner of expanding the CT. Nevertheless, it can be observed that the runtime increases exponentially in general with respect to the number of bombs per subtask in Figure~\ref{fig:plots_num_bombs_per_subtask}. This is expected given the combinatorial enumeration of task assignment to OPEN before any CT expansion as shown in line~\ref{alg:cbs_ta_ptc_combinations_start}--\ref{alg:cbs_ta_ptc_combinations_end} of Algorithm~\ref{alg:cbs_ta_ptc}.

From Figure~\ref{fig:plots_seconds_per_timestep}, the success rate and optimality ratio naturally decrease with an increasing seconds per timestep parameter for both CBS-TA-PTC and CBS-TA given the growing difficulty of the task with shorter bomb fuses and countdowns in terms of timesteps. By calculating the average inter-node distance and using the default walking speed of $4.317 \: ms^{-1}$ in Minecraft, we find that $2$ seconds per timestep is the best approximation of the Minecraft dynamics, to which we have a success rate of approximately $0.65$ for CBS-TA-PTC compared to $0.15$ for CBS-TA. Hence, it is clear that the usage of vertex and edge conflicts alone in CBS-TA is insufficient to solve TAPF-PTC problems efficiently in general. 

\section{Discussion}
\label{sec:discussion}

As CBS-TA-PTC is an incomplete and suboptimal algorithm by 
decomposing the task into subtasks using user-defined heuristics, we discuss guidelines to select good heuristics to sort the order of the goals for the entire task as shown in line~\ref{alg:cbs_ta_ptc_task_partition_heuristic_sort} in Algorithm~\ref{alg:cbs_ta_ptc_task_partition}. In general, a good criterion would be to select a heuristic based on attributes that determine absolute temporal range constraints, given that they are hard constraints that are the natural first order of priority. In the case of the bomb-defusing environment, the attribute that determines absolute temporal range constraints is the bomb fuse length. Hence, the task allocation heuristic used sorts the bombs in ascending order of fuse length. Returning to the automated food service context as a real-world motivation mentioned in Section~\ref{sec:problem_definition}, the absolute temporal range constraint is determined by the order in which customers arrive, leading to a practical first-come-first-serve heuristic. In cases of tasks without absolute temporal range constraints, a sensible heuristic would be to use the K-Means clustering algorithm \cite{hartigan1979algorithm}, where
$k = \text{number of goals per subtask}//\text{number of agents}$, to cluster goals based on their locations and arrange them such that agents would visit the nearest cluster from their previous location.

As highlighted in Section~\ref{subsubsec:experiments_results_cbs}, a key limitation of CBS-TA-PTC is its combinatorial enumeration of task assignment to OPEN before any CT expansion as shown in line~\ref{alg:cbs_ta_ptc_combinations_start}--\ref{alg:cbs_ta_ptc_combinations_end} of Algorithm~\ref{alg:cbs_ta_ptc} that causes its runtime to scale exponentially to the number of bombs per subtask. The underlying motivation is to ensure that any root node inserted to OPEN to be expanded has the highest return possible by ordering the entire task assignment based on return as shown in line~\ref{alg:cbs_ta_ptc_open_highest_return} in Algorithm~\ref{alg:cbs_ta_ptc}. Hence, the corresponding CT generated from the root nodes would ideally have as few conflicts as possible before the optimal solution is found to help reduce runtime. Furthermore, it is possible to chance upon an optimal conflict-free solution during the enumeration as shown in line~\ref{alg:cbs_ta_ptc_root_maximum_return_start}--\ref{alg:cbs_ta_ptc_root_maximum_return_end} in Algorithm~\ref{alg:cbs_ta_ptc}. A consequence of this design choice for CBS-TA-PTC is that it pushes the combinatorial enumeration of task assignment as the key contributor to the runtime rather than the conflict resolution process shown in line~\ref{alg:cbs_ta_ptc_while_open_start}--\ref{alg:cbs_ta_ptc_while_open_end} of Algorithm~\ref{alg:cbs_ta_ptc}. 

To address the above issue, an option would be to design a variant of CBS-TA-PTC where the task assignment is generated on demand similar to the high level of CBS-TA~\cite{honig2018conflict}. However, unlike CBS-TA where the next task assignment can be generated using any optimal assignment algorithm (e.g. Hungarian method~\cite{kuhn1955hungarian}), the next best task assignment that maximizes the return is generally unknown. Therefore, a random task assignment that has yet to be generated would be used, unless task-specific domain knowledge can be utilized to generate a bias towards specific types of task assignment. For example, a simple heuristic that can be applied to both variants of CBS methods would be to favor task assignments where the number of goals assigned to each agent is approximately equal. Naturally, a major drawback would be the increase in runtime by expanding the CT of unfeasible task assignments in terms of optimality. Hence, the better alternative would be to parallelize the generation of the root nodes to OPEN from combinations of task assignment by leveraging the usage of multiple parallel oracles that are typically implemented as a user-designed RL environment based on the Gymnasium API \cite{towers_gymnasium_2023} to evaluate the return of a given task assignment as shown in line~\ref{alg:cbs_ta_ptc_root_return} of Algorithm~\ref{alg:cbs_ta_ptc}, a routine that consumes the most runtime for the task assignment enumeration. Given modern computing resources, the parallelization can be implemented practically and efficiently and would greatly ameliorate the runtime drawback of the CBS-TA-PTC from the combinatorial enumeration of task assignment, making CBS-TA-PTC highly viable for real-world applications.

Lastly, it is important to note that CBS-TA-PTC is unable to address stochasticity in environments, which are best addressed with learning methods that can potentially generalize to novel and stochastic events. Despite the failure of MARL methods for the bomb-defusing environment, future work could involve the usage of CBS-TA-PTC as an expert policy to generate trajectories to assist MARL methods via imitation learning in a similar manner as shown in PRIMAL~\cite{sartoretti2019primal}, along with methods from curriculum learning (e.g. scaling difficulty of the task as training progresses). We believe that the stated training pipeline can greatly help the MARL methods become more sample-efficient in learning and hence potentially more viable and practical for applications in real-world scenarios and tasks.

\section{Conclusion}
\label{sec:conclusion}

In this paper, we propose the CBS-TA-PTC algorithm that simultaneously generates task assignments as well as feasible and collision-free paths that adhere to precedence and temporal constraints for agents to maximize a user-defined objective to solve real-world problems. We evaluate CBS-TA-PTC in a highly challenging bomb-defusing environment with precedence and temporal constraints and show that it can efficiently solve tasks relative to MARL and adapted TAPF methods.



\begin{acks}
This work was partially supported by DARPA award HR001120C0036 and AFOSR award FA9550-18-1-0097. We thank Inioluwa Oguntola for his contributions to the development of the environment for the bomb-defusing task.
\end{acks}



\balance
\bibliographystyle{ACM-Reference-Format} 
\bibliography{references}


\begin{thebibliography}{33}


\ifx \showCODEN    \undefined \def \showCODEN     #1{\unskip}     \fi
\ifx \showDOI      \undefined \def \showDOI       #1{#1}\fi
\ifx \showISBNx    \undefined \def \showISBNx     #1{\unskip}     \fi
\ifx \showISBNxiii \undefined \def \showISBNxiii  #1{\unskip}     \fi
\ifx \showISSN     \undefined \def \showISSN      #1{\unskip}     \fi
\ifx \showLCCN     \undefined \def \showLCCN      #1{\unskip}     \fi
\ifx \shownote     \undefined \def \shownote      #1{#1}          \fi
\ifx \showarticletitle \undefined \def \showarticletitle #1{#1}   \fi
\ifx \showURL      \undefined \def \showURL       {\relax}        \fi
\providecommand\bibfield[2]{#2}
\providecommand\bibinfo[2]{#2}
\providecommand\natexlab[1]{#1}
\providecommand\showeprint[2][]{arXiv:#2}

\bibitem[\protect\citeauthoryear{Bard, Foerster, Chandar, Burch, Lanctot, Song, Parisotto, Dumoulin, Moitra, Hughes, et~al\mbox{.}}{Bard et~al\mbox{.}}{2020}]%
        {bard2020hanabi}
\bibfield{author}{\bibinfo{person}{Nolan Bard}, \bibinfo{person}{Jakob~N Foerster}, \bibinfo{person}{Sarath Chandar}, \bibinfo{person}{Neil Burch}, \bibinfo{person}{Marc Lanctot}, \bibinfo{person}{H~Francis Song}, \bibinfo{person}{Emilio Parisotto}, \bibinfo{person}{Vincent Dumoulin}, \bibinfo{person}{Subhodeep Moitra}, \bibinfo{person}{Edward Hughes}, {et~al\mbox{.}}} \bibinfo{year}{2020}\natexlab{}.
\newblock \showarticletitle{The hanabi challenge: A new frontier for ai research}.
\newblock \bibinfo{journal}{\emph{Artificial Intelligence}}  \bibinfo{volume}{280} (\bibinfo{year}{2020}), \bibinfo{pages}{103216}.
\newblock


\bibitem[\protect\citeauthoryear{Chen, Alonso-Mora, Bai, Harabor, and Stuckey}{Chen et~al\mbox{.}}{2021}]%
        {chen2021integrated}
\bibfield{author}{\bibinfo{person}{Zhe Chen}, \bibinfo{person}{Javier Alonso-Mora}, \bibinfo{person}{Xiaoshan Bai}, \bibinfo{person}{Daniel~D Harabor}, {and} \bibinfo{person}{Peter~J Stuckey}.} \bibinfo{year}{2021}\natexlab{}.
\newblock \showarticletitle{Integrated task assignment and path planning for capacitated multi-agent pickup and delivery}.
\newblock \bibinfo{journal}{\emph{IEEE Robotics and Automation Letters}} \bibinfo{volume}{6}, \bibinfo{number}{3} (\bibinfo{year}{2021}), \bibinfo{pages}{5816--5823}.
\newblock


\bibitem[\protect\citeauthoryear{Ferner, Wagner, and Choset}{Ferner et~al\mbox{.}}{2013}]%
        {ferner2013odrm}
\bibfield{author}{\bibinfo{person}{Cornelia Ferner}, \bibinfo{person}{Glenn Wagner}, {and} \bibinfo{person}{Howie Choset}.} \bibinfo{year}{2013}\natexlab{}.
\newblock \showarticletitle{ODrM* optimal multirobot path planning in low dimensional search spaces}. In \bibinfo{booktitle}{\emph{2013 IEEE International Conference on Robotics and Automation}}. IEEE, \bibinfo{pages}{3854--3859}.
\newblock


\bibitem[\protect\citeauthoryear{Goldenberg, Felner, Stern, Sharon, Sturtevant, Holte, and Schaeffer}{Goldenberg et~al\mbox{.}}{2014}]%
        {goldenberg2014enhanced}
\bibfield{author}{\bibinfo{person}{Meir Goldenberg}, \bibinfo{person}{Ariel Felner}, \bibinfo{person}{Roni Stern}, \bibinfo{person}{Guni Sharon}, \bibinfo{person}{Nathan Sturtevant}, \bibinfo{person}{Robert~C Holte}, {and} \bibinfo{person}{Jonathan Schaeffer}.} \bibinfo{year}{2014}\natexlab{}.
\newblock \showarticletitle{Enhanced partial expansion a}.
\newblock \bibinfo{journal}{\emph{Journal of Artificial Intelligence Research}}  \bibinfo{volume}{50} (\bibinfo{year}{2014}), \bibinfo{pages}{141--187}.
\newblock


\bibitem[\protect\citeauthoryear{Grenouilleau, van Hoeve, and Hooker}{Grenouilleau et~al\mbox{.}}{2019}]%
        {grenouilleau2019multi}
\bibfield{author}{\bibinfo{person}{Florian Grenouilleau}, \bibinfo{person}{Willem-Jan van Hoeve}, {and} \bibinfo{person}{John~N Hooker}.} \bibinfo{year}{2019}\natexlab{}.
\newblock \showarticletitle{A multi-label A* algorithm for multi-agent pathfinding}. In \bibinfo{booktitle}{\emph{Proceedings of the International Conference on Automated Planning and Scheduling}}, Vol.~\bibinfo{volume}{29}. \bibinfo{pages}{181--185}.
\newblock


\bibitem[\protect\citeauthoryear{Hall, Galabova, Gottwald, and Feldmeier}{Hall et~al\mbox{.}}{[n.d.]}]%
        {hallhighs}
\bibfield{author}{\bibinfo{person}{J Hall}, \bibinfo{person}{I Galabova}, \bibinfo{person}{L Gottwald}, {and} \bibinfo{person}{M Feldmeier}.} \bibinfo{year}{[n.d.]}\natexlab{}.
\newblock \bibinfo{title}{HiGHS--high performance software for linear optimization}.
\newblock
\newblock


\bibitem[\protect\citeauthoryear{Hartigan and Wong}{Hartigan and Wong}{1979}]%
        {hartigan1979algorithm}
\bibfield{author}{\bibinfo{person}{John~A Hartigan} {and} \bibinfo{person}{Manchek~A Wong}.} \bibinfo{year}{1979}\natexlab{}.
\newblock \showarticletitle{Algorithm AS 136: A k-means clustering algorithm}.
\newblock \bibinfo{journal}{\emph{Journal of the royal statistical society. series c (applied statistics)}} \bibinfo{volume}{28}, \bibinfo{number}{1} (\bibinfo{year}{1979}), \bibinfo{pages}{100--108}.
\newblock


\bibitem[\protect\citeauthoryear{H{\"o}nig, Kiesel, Tinka, Durham, and Ayanian}{H{\"o}nig et~al\mbox{.}}{2018}]%
        {honig2018conflict}
\bibfield{author}{\bibinfo{person}{Wolfgang H{\"o}nig}, \bibinfo{person}{Scott Kiesel}, \bibinfo{person}{Andrew Tinka}, \bibinfo{person}{Joseph Durham}, {and} \bibinfo{person}{Nora Ayanian}.} \bibinfo{year}{2018}\natexlab{}.
\newblock \showarticletitle{Conflict-based search with optimal task assignment}. In \bibinfo{booktitle}{\emph{Proceedings of the International Joint Conference on Autonomous Agents and Multiagent Systems}}.
\newblock


\bibitem[\protect\citeauthoryear{Huangfu and Hall}{Huangfu and Hall}{2018}]%
        {huangfu2018parallelizing}
\bibfield{author}{\bibinfo{person}{Qi Huangfu} {and} \bibinfo{person}{JA~Julian Hall}.} \bibinfo{year}{2018}\natexlab{}.
\newblock \showarticletitle{Parallelizing the dual revised simplex method}.
\newblock \bibinfo{journal}{\emph{Mathematical Programming Computation}} \bibinfo{volume}{10}, \bibinfo{number}{1} (\bibinfo{year}{2018}), \bibinfo{pages}{119--142}.
\newblock


\bibitem[\protect\citeauthoryear{Kuhn}{Kuhn}{1955}]%
        {kuhn1955hungarian}
\bibfield{author}{\bibinfo{person}{Harold~W Kuhn}.} \bibinfo{year}{1955}\natexlab{}.
\newblock \showarticletitle{The Hungarian method for the assignment problem}.
\newblock \bibinfo{journal}{\emph{Naval research logistics quarterly}} \bibinfo{volume}{2}, \bibinfo{number}{1-2} (\bibinfo{year}{1955}), \bibinfo{pages}{83--97}.
\newblock


\bibitem[\protect\citeauthoryear{LaValle}{LaValle}{2006}]%
        {lavalle2006planning}
\bibfield{author}{\bibinfo{person}{Steven~M LaValle}.} \bibinfo{year}{2006}\natexlab{}.
\newblock \bibinfo{booktitle}{\emph{Planning algorithms}}.
\newblock \bibinfo{publisher}{Cambridge university press}.
\newblock


\bibitem[\protect\citeauthoryear{Littman}{Littman}{1994}]%
        {littman1994markov}
\bibfield{author}{\bibinfo{person}{Michael~L Littman}.} \bibinfo{year}{1994}\natexlab{}.
\newblock \showarticletitle{Markov games as a framework for multi-agent reinforcement learning}.
\newblock In \bibinfo{booktitle}{\emph{Machine learning proceedings 1994}}. \bibinfo{publisher}{Elsevier}, \bibinfo{pages}{157--163}.
\newblock


\bibitem[\protect\citeauthoryear{Lowe, Wu, Tamar, Harb, Abbeel, and Mordatch}{Lowe et~al\mbox{.}}{2017}]%
        {lowe2017multi}
\bibfield{author}{\bibinfo{person}{Ryan Lowe}, \bibinfo{person}{Yi Wu}, \bibinfo{person}{Aviv Tamar}, \bibinfo{person}{Jean Harb}, \bibinfo{person}{Pieter Abbeel}, {and} \bibinfo{person}{Igor Mordatch}.} \bibinfo{year}{2017}\natexlab{}.
\newblock \showarticletitle{Multi-Agent Actor-Critic for Mixed Cooperative-Competitive Environments}.
\newblock \bibinfo{journal}{\emph{Neural Information Processing Systems (NIPS)}} (\bibinfo{year}{2017}).
\newblock


\bibitem[\protect\citeauthoryear{Ma, Harabor, Stuckey, Li, and Koenig}{Ma et~al\mbox{.}}{2019}]%
        {ma2019searching}
\bibfield{author}{\bibinfo{person}{Hang Ma}, \bibinfo{person}{Daniel Harabor}, \bibinfo{person}{Peter~J Stuckey}, \bibinfo{person}{Jiaoyang Li}, {and} \bibinfo{person}{Sven Koenig}.} \bibinfo{year}{2019}\natexlab{}.
\newblock \showarticletitle{Searching with consistent prioritization for multi-agent path finding}. In \bibinfo{booktitle}{\emph{Proceedings of the AAAI Conference on Artificial Intelligence}}, Vol.~\bibinfo{volume}{33}. \bibinfo{pages}{7643--7650}.
\newblock


\bibitem[\protect\citeauthoryear{Ma and Koenig}{Ma and Koenig}{2016}]%
        {ma2016optimal}
\bibfield{author}{\bibinfo{person}{Hang Ma} {and} \bibinfo{person}{Sven Koenig}.} \bibinfo{year}{2016}\natexlab{}.
\newblock \showarticletitle{Optimal target assignment and path finding for teams of agents}.
\newblock \bibinfo{journal}{\emph{arXiv preprint arXiv:1612.05693}} (\bibinfo{year}{2016}).
\newblock


\bibitem[\protect\citeauthoryear{Ma, Tovey, Sharon, Kumar, and Koenig}{Ma et~al\mbox{.}}{2016}]%
        {ma2016multi}
\bibfield{author}{\bibinfo{person}{Hang Ma}, \bibinfo{person}{Craig Tovey}, \bibinfo{person}{Guni Sharon}, \bibinfo{person}{TK Kumar}, {and} \bibinfo{person}{Sven Koenig}.} \bibinfo{year}{2016}\natexlab{}.
\newblock \showarticletitle{Multi-agent path finding with payload transfers and the package-exchange robot-routing problem}. In \bibinfo{booktitle}{\emph{Proceedings of the AAAI Conference on Artificial Intelligence}}, Vol.~\bibinfo{volume}{30}.
\newblock


\bibitem[\protect\citeauthoryear{{Mojang Studios}}{{Mojang Studios}}{2011}]%
        {minecraft}
\bibfield{author}{\bibinfo{person}{{Mojang Studios}}.} \bibinfo{year}{2011}\natexlab{}.
\newblock \bibinfo{title}{Minecraft}.
\newblock \bibinfo{howpublished}{\url{https://www.minecraft.net/}}.
\newblock


\bibitem[\protect\citeauthoryear{Mordatch and Abbeel}{Mordatch and Abbeel}{2017}]%
        {mordatch2017emergence}
\bibfield{author}{\bibinfo{person}{Igor Mordatch} {and} \bibinfo{person}{Pieter Abbeel}.} \bibinfo{year}{2017}\natexlab{}.
\newblock \showarticletitle{Emergence of Grounded Compositional Language in Multi-Agent Populations}.
\newblock \bibinfo{journal}{\emph{arXiv preprint arXiv:1703.04908}} (\bibinfo{year}{2017}).
\newblock


\bibitem[\protect\citeauthoryear{Morris, Pasareanu, Luckow, Malik, Ma, Kumar, and Koenig}{Morris et~al\mbox{.}}{2016}]%
        {morris2016planning}
\bibfield{author}{\bibinfo{person}{Robert Morris}, \bibinfo{person}{Corina~S Pasareanu}, \bibinfo{person}{Kasper~S{\o}e Luckow}, \bibinfo{person}{Waqar Malik}, \bibinfo{person}{Hang Ma}, \bibinfo{person}{TK~Satish Kumar}, {and} \bibinfo{person}{Sven Koenig}.} \bibinfo{year}{2016}\natexlab{}.
\newblock \showarticletitle{Planning, Scheduling and Monitoring for Airport Surface Operations.}. In \bibinfo{booktitle}{\emph{AAAI Workshop: Planning for Hybrid Systems}}. \bibinfo{pages}{608--614}.
\newblock


\bibitem[\protect\citeauthoryear{Okumura and D{\'e}fago}{Okumura and D{\'e}fago}{2023}]%
        {okumura2023solving}
\bibfield{author}{\bibinfo{person}{Keisuke Okumura} {and} \bibinfo{person}{Xavier D{\'e}fago}.} \bibinfo{year}{2023}\natexlab{}.
\newblock \showarticletitle{Solving simultaneous target assignment and path planning efficiently with time-independent execution}.
\newblock \bibinfo{journal}{\emph{Artificial Intelligence}}  \bibinfo{volume}{321} (\bibinfo{year}{2023}), \bibinfo{pages}{103946}.
\newblock


\bibitem[\protect\citeauthoryear{Rashid, Samvelyan, De~Witt, Farquhar, Foerster, and Whiteson}{Rashid et~al\mbox{.}}{2020}]%
        {rashid2020monotonic}
\bibfield{author}{\bibinfo{person}{Tabish Rashid}, \bibinfo{person}{Mikayel Samvelyan}, \bibinfo{person}{Christian~Schroeder De~Witt}, \bibinfo{person}{Gregory Farquhar}, \bibinfo{person}{Jakob Foerster}, {and} \bibinfo{person}{Shimon Whiteson}.} \bibinfo{year}{2020}\natexlab{}.
\newblock \showarticletitle{Monotonic value function factorisation for deep multi-agent reinforcement learning}.
\newblock \bibinfo{journal}{\emph{The Journal of Machine Learning Research}} \bibinfo{volume}{21}, \bibinfo{number}{1} (\bibinfo{year}{2020}), \bibinfo{pages}{7234--7284}.
\newblock


\bibitem[\protect\citeauthoryear{Samvelyan, Rashid, de~Witt, Farquhar, Nardelli, Rudner, Hung, Torr, Foerster, and Whiteson}{Samvelyan et~al\mbox{.}}{2019}]%
        {samvelyan19smac}
\bibfield{author}{\bibinfo{person}{Mikayel Samvelyan}, \bibinfo{person}{Tabish Rashid}, \bibinfo{person}{Christian~Schroeder de Witt}, \bibinfo{person}{Gregory Farquhar}, \bibinfo{person}{Nantas Nardelli}, \bibinfo{person}{Tim G.~J. Rudner}, \bibinfo{person}{Chia-Man Hung}, \bibinfo{person}{Philiph H.~S. Torr}, \bibinfo{person}{Jakob Foerster}, {and} \bibinfo{person}{Shimon Whiteson}.} \bibinfo{year}{2019}\natexlab{}.
\newblock \showarticletitle{{The} {StarCraft} {Multi}-{Agent} {Challenge}}.
\newblock \bibinfo{journal}{\emph{CoRR}}  \bibinfo{volume}{abs/1902.04043} (\bibinfo{year}{2019}).
\newblock


\bibitem[\protect\citeauthoryear{Sartoretti, Kerr, Shi, Wagner, Kumar, Koenig, and Choset}{Sartoretti et~al\mbox{.}}{2019}]%
        {sartoretti2019primal}
\bibfield{author}{\bibinfo{person}{Guillaume Sartoretti}, \bibinfo{person}{Justin Kerr}, \bibinfo{person}{Yunfei Shi}, \bibinfo{person}{Glenn Wagner}, \bibinfo{person}{TK~Satish Kumar}, \bibinfo{person}{Sven Koenig}, {and} \bibinfo{person}{Howie Choset}.} \bibinfo{year}{2019}\natexlab{}.
\newblock \showarticletitle{Primal: Pathfinding via reinforcement and imitation multi-agent learning}.
\newblock \bibinfo{journal}{\emph{IEEE Robotics and Automation Letters}} \bibinfo{volume}{4}, \bibinfo{number}{3} (\bibinfo{year}{2019}), \bibinfo{pages}{2378--2385}.
\newblock


\bibitem[\protect\citeauthoryear{Sharon, Stern, Felner, and Sturtevant}{Sharon et~al\mbox{.}}{2015}]%
        {sharon2015conflict}
\bibfield{author}{\bibinfo{person}{Guni Sharon}, \bibinfo{person}{Roni Stern}, \bibinfo{person}{Ariel Felner}, {and} \bibinfo{person}{Nathan~R Sturtevant}.} \bibinfo{year}{2015}\natexlab{}.
\newblock \showarticletitle{Conflict-based search for optimal multi-agent pathfinding}.
\newblock \bibinfo{journal}{\emph{Artificial Intelligence}}  \bibinfo{volume}{219} (\bibinfo{year}{2015}), \bibinfo{pages}{40--66}.
\newblock


\bibitem[\protect\citeauthoryear{Shaw}{Shaw}{1998}]%
        {shaw1998using}
\bibfield{author}{\bibinfo{person}{Paul Shaw}.} \bibinfo{year}{1998}\natexlab{}.
\newblock \showarticletitle{Using constraint programming and local search methods to solve vehicle routing problems}. In \bibinfo{booktitle}{\emph{International conference on principles and practice of constraint programming}}. Springer, \bibinfo{pages}{417--431}.
\newblock


\bibitem[\protect\citeauthoryear{Silver}{Silver}{2005}]%
        {silver2005cooperative}
\bibfield{author}{\bibinfo{person}{David Silver}.} \bibinfo{year}{2005}\natexlab{}.
\newblock \showarticletitle{Cooperative pathfinding}. In \bibinfo{booktitle}{\emph{Proceedings of the aaai conference on artificial intelligence and interactive digital entertainment}}, Vol.~\bibinfo{volume}{1}. \bibinfo{pages}{117--122}.
\newblock


\bibitem[\protect\citeauthoryear{Stern, Sturtevant, Felner, Koenig, Ma, Walker, Li, Atzmon, Cohen, Kumar, et~al\mbox{.}}{Stern et~al\mbox{.}}{2019}]%
        {stern2019multi}
\bibfield{author}{\bibinfo{person}{Roni Stern}, \bibinfo{person}{Nathan Sturtevant}, \bibinfo{person}{Ariel Felner}, \bibinfo{person}{Sven Koenig}, \bibinfo{person}{Hang Ma}, \bibinfo{person}{Thayne Walker}, \bibinfo{person}{Jiaoyang Li}, \bibinfo{person}{Dor Atzmon}, \bibinfo{person}{Liron Cohen}, \bibinfo{person}{TK Kumar}, {et~al\mbox{.}}} \bibinfo{year}{2019}\natexlab{}.
\newblock \showarticletitle{Multi-agent pathfinding: Definitions, variants, and benchmarks}. In \bibinfo{booktitle}{\emph{Proceedings of the International Symposium on Combinatorial Search}}, Vol.~\bibinfo{volume}{10}. \bibinfo{pages}{151--158}.
\newblock


\bibitem[\protect\citeauthoryear{Towers, Terry, Kwiatkowski, Balis, Cola, Deleu, Goulão, Kallinteris, KG, Krimmel, Perez-Vicente, Pierré, Schulhoff, Tai, Shen, and Younis}{Towers et~al\mbox{.}}{2023}]%
        {towers_gymnasium_2023}
\bibfield{author}{\bibinfo{person}{Mark Towers}, \bibinfo{person}{Jordan~K. Terry}, \bibinfo{person}{Ariel Kwiatkowski}, \bibinfo{person}{John~U. Balis}, \bibinfo{person}{Gianluca~de Cola}, \bibinfo{person}{Tristan Deleu}, \bibinfo{person}{Manuel Goulão}, \bibinfo{person}{Andreas Kallinteris}, \bibinfo{person}{Arjun KG}, \bibinfo{person}{Markus Krimmel}, \bibinfo{person}{Rodrigo Perez-Vicente}, \bibinfo{person}{Andrea Pierré}, \bibinfo{person}{Sander Schulhoff}, \bibinfo{person}{Jun~Jet Tai}, \bibinfo{person}{Andrew Tan~Jin Shen}, {and} \bibinfo{person}{Omar~G. Younis}.} \bibinfo{year}{2023}\natexlab{}.
\newblock \bibinfo{title}{Gymnasium}.
\newblock
\newblock
\urldef\tempurl%
\url{https://doi.org/10.5281/zenodo.8127026}
\showDOI{\tempurl}


\bibitem[\protect\citeauthoryear{Wagner and Choset}{Wagner and Choset}{2015}]%
        {wagner2015subdimensional}
\bibfield{author}{\bibinfo{person}{Glenn Wagner} {and} \bibinfo{person}{Howie Choset}.} \bibinfo{year}{2015}\natexlab{}.
\newblock \showarticletitle{Subdimensional expansion for multirobot path planning}.
\newblock \bibinfo{journal}{\emph{Artificial intelligence}}  \bibinfo{volume}{219} (\bibinfo{year}{2015}), \bibinfo{pages}{1--24}.
\newblock


\bibitem[\protect\citeauthoryear{Wurman, D'Andrea, and Mountz}{Wurman et~al\mbox{.}}{2008}]%
        {wurman2008coordinating}
\bibfield{author}{\bibinfo{person}{Peter~R Wurman}, \bibinfo{person}{Raffaello D'Andrea}, {and} \bibinfo{person}{Mick Mountz}.} \bibinfo{year}{2008}\natexlab{}.
\newblock \showarticletitle{Coordinating hundreds of cooperative, autonomous vehicles in warehouses}.
\newblock \bibinfo{journal}{\emph{AI magazine}} \bibinfo{volume}{29}, \bibinfo{number}{1} (\bibinfo{year}{2008}), \bibinfo{pages}{9--9}.
\newblock


\bibitem[\protect\citeauthoryear{Yu, Velu, Vinitsky, Gao, Wang, Bayen, and Wu}{Yu et~al\mbox{.}}{2022}]%
        {yu2022surprising}
\bibfield{author}{\bibinfo{person}{Chao Yu}, \bibinfo{person}{Akash Velu}, \bibinfo{person}{Eugene Vinitsky}, \bibinfo{person}{Jiaxuan Gao}, \bibinfo{person}{Yu Wang}, \bibinfo{person}{Alexandre Bayen}, {and} \bibinfo{person}{Yi Wu}.} \bibinfo{year}{2022}\natexlab{}.
\newblock \showarticletitle{The surprising effectiveness of ppo in cooperative multi-agent games}.
\newblock \bibinfo{journal}{\emph{Advances in Neural Information Processing Systems}}  \bibinfo{volume}{35} (\bibinfo{year}{2022}), \bibinfo{pages}{24611--24624}.
\newblock


\bibitem[\protect\citeauthoryear{Yu and Rus}{Yu and Rus}{2015}]%
        {yu2015pebble}
\bibfield{author}{\bibinfo{person}{Jingjin Yu} {and} \bibinfo{person}{Daniela Rus}.} \bibinfo{year}{2015}\natexlab{}.
\newblock \showarticletitle{Pebble motion on graphs with rotations: Efficient feasibility tests and planning algorithms}. In \bibinfo{booktitle}{\emph{Algorithmic Foundations of Robotics XI: Selected Contributions of the Eleventh International Workshop on the Algorithmic Foundations of Robotics}}. Springer, \bibinfo{pages}{729--746}.
\newblock


\bibitem[\protect\citeauthoryear{Zhang, Chen, Li, Williams, and Koenig}{Zhang et~al\mbox{.}}{2022}]%
        {zhang2022multi}
\bibfield{author}{\bibinfo{person}{Han Zhang}, \bibinfo{person}{Jingkai Chen}, \bibinfo{person}{Jiaoyang Li}, \bibinfo{person}{B Williams}, {and} \bibinfo{person}{Sven Koenig}.} \bibinfo{year}{2022}\natexlab{}.
\newblock \showarticletitle{Multi-Agent Path Finding for Precedence-Constrained Goal Sequences}. In \bibinfo{booktitle}{\emph{Proceedings of the International Joint Conference on Autonomous Agents and Multiagent Systems}}. \bibinfo{pages}{1464–--1472}.
\newblock


\end{thebibliography}


\end{document}